\documentclass[runningheads]{llncs}
\usepackage[T1]{fontenc}
\usepackage{graphicx}
\usepackage{verbatim}
\usepackage{times}
\usepackage{epsfig}
\usepackage{graphicx}
\usepackage{amsmath}
\usepackage{amssymb}

\usepackage{foreign}
\usepackage{cite}
\usepackage{tablefootnote}
\usepackage{booktabs}
\usepackage{xcolor}
\usepackage[normalem]{ulem}
\usepackage[flushleft]{threeparttable}
\newcommand{\ie}{\textit{i}.\textit{e}.}
\newcommand{\eg}{\textit{e}.\textit{g}.}

\begin{document}
\title{PIFu for the Real World: A Self-supervised Framework to Reconstruct Dressed Human from Single-view Images}
%
%
\author{Zhangyang Xiong\inst{1,2} \and Dong Du\inst{2} \and Yushuang Wu\inst{2} \and Jingqi Dong\inst{2} \and Di Kang\inst{3} \and \\ Linchao Bao\inst{3} \and
 Xiaoguang Han\inst{2,1*}
}
%
\authorrunning{Z. Xiong et al.}
%

\institute{FNii, The Chinese University of Hong Kong-Shenzhen, Shenzhen, China \and SSE, The Chinese University of Hong Kong-Shenzhen, Shenzhen, China \and Tencent AI Lab, Shenzhen, China}
%
\maketitle              
\begin{abstract}
It is very challenging to accurately reconstruct sophisticated human geometry caused by various poses and garments from a single image.
Recently, works based on pixel-aligned implicit function (PIFu) have made a big step and achieved state-of-the-art fidelity on image-based 3D human digitization.
However, the training of PIFu relies heavily on expensive and limited 3D ground truth data (i.e. synthetic data), thus hindering its generalization to more diverse real world images.
In this work, we propose an end-to-end self-supervised network named SelfPIFu to utilize abundant and diverse in-the-wild images,resulting in largely improved reconstructions when tested on unconstrained in-the-wild images.
At the core of SelfPIFu is the depth-guided volume-/surface-aware signed distance fields (SDF) learning, which enables self-supervised learning of a PIFu without access to GT mesh.
The whole framework consists of a normal estimator, a depth estimator, and a SDF-based PIFu and better utilizes extra depth GT during training.
Extensive experiments demonstrate the effectiveness of our self-supervised framework and the superiority of using depth as input. 
On synthetic data, our Intersection-Over-Union (IoU) achieves to 89.03\%, 20\% and 28.6\%  higher compared with PIFuHD and ECON, respectively.
For in-the-wild images, our method excels at reconstructing geometric details that are both rich and highly representative of the actual human, as illustrated in Fig.~\ref{fig_teaser} and~\ref{fig:compare_real}.
\end{abstract}
\section{Introduction}
\label{sec:introduction}

Image-based human digitization has gained considerable attention in the last decades. 
It is widely used in games, telepresence, and VR/AR applications. 
To recover accurate 3D human shapes from sparse 2D observations, various models are proposed, such as parametric models~\cite{anguelov2005scape,loper2015smpl,pons2015dyna,hmrKanazawa17,lassner2017unite,joo2018total,omran2018neural,pavlakos2019expressive}, 
 silhouetted models~\cite{natsume2019siclope}, volumetric models~\cite{varol2018bodynet,zheng2019deephuman}, and implicit models~\cite{chibane2020implicit,saito2019pifu,saito2020pifuhd}. 
Among them, PIFuHD~\cite{saito2020pifuhd} produces high-fidelity 3D reconstruction with impressive geometric details such as wrinkles of clothes, which achieves the state-of-the-art.

Compared with the original version PIFu~\cite{saito2019pifu}, PIFuHD~\cite{saito2020pifuhd} proposes a two-level pixel-aligned implicit function learning framework for high-resolution 3D reconstructions, 
where additional normal maps are generated and integrated for fine-grained detail preservation. 
However, PIFuHD still suffers from two issues: 
(i) PIFuHD relies heavily on delicately created 3D ground truth supervision and normal map, while the existing dataset, \eg~RenderPeople\cite{renerpeople}, contains only a few hundred static models covering limited identities, poses, and complex garment geometry, resulting in performance degradation when applied on in-the-wild images; 
(ii) the image-to-normal-to-shape reconstruction of PIFuHD is constructed with normal maps as the intermediate while the normal estimation is sensitive to texture and shadow in the input image~\cite{tang2019neural,wang2020normalgan,jafarian2021learning}.

Considering the limited synthetic normal data cannot cover all varieties of cloth textures and complex shadows in the real world, estimating normal maps from real world images is usually difficult and error-prone.
The implicit function may not handle these errors in normal map which lead to artifacts in shape reconstruction quality, as illustrated in Fig.~\ref{fig:vis_NvsD} and ~\ref{fig:compare_real}.
Thus, the heavy reliance on 3D ground truth and the poor robustness of normal estimation of PIFuHD motivates us to rethink the whole reconstruction procedure. 
Compared to using normal maps, depth maps, which contain not only rich and detailed shape information but also \emph{directly} define the surface,are able to provide spatial supervision beyond the surface and make it possible to learn a more promising signed distance field (SDF).
What's more, depth estimators generalize well from synthetic data to in-the-wild images
(see Fig.~\ref{fig:compare_dn_details}). 

\begin{figure*}[t]
    \centering
    \includegraphics[width=0.99\linewidth]{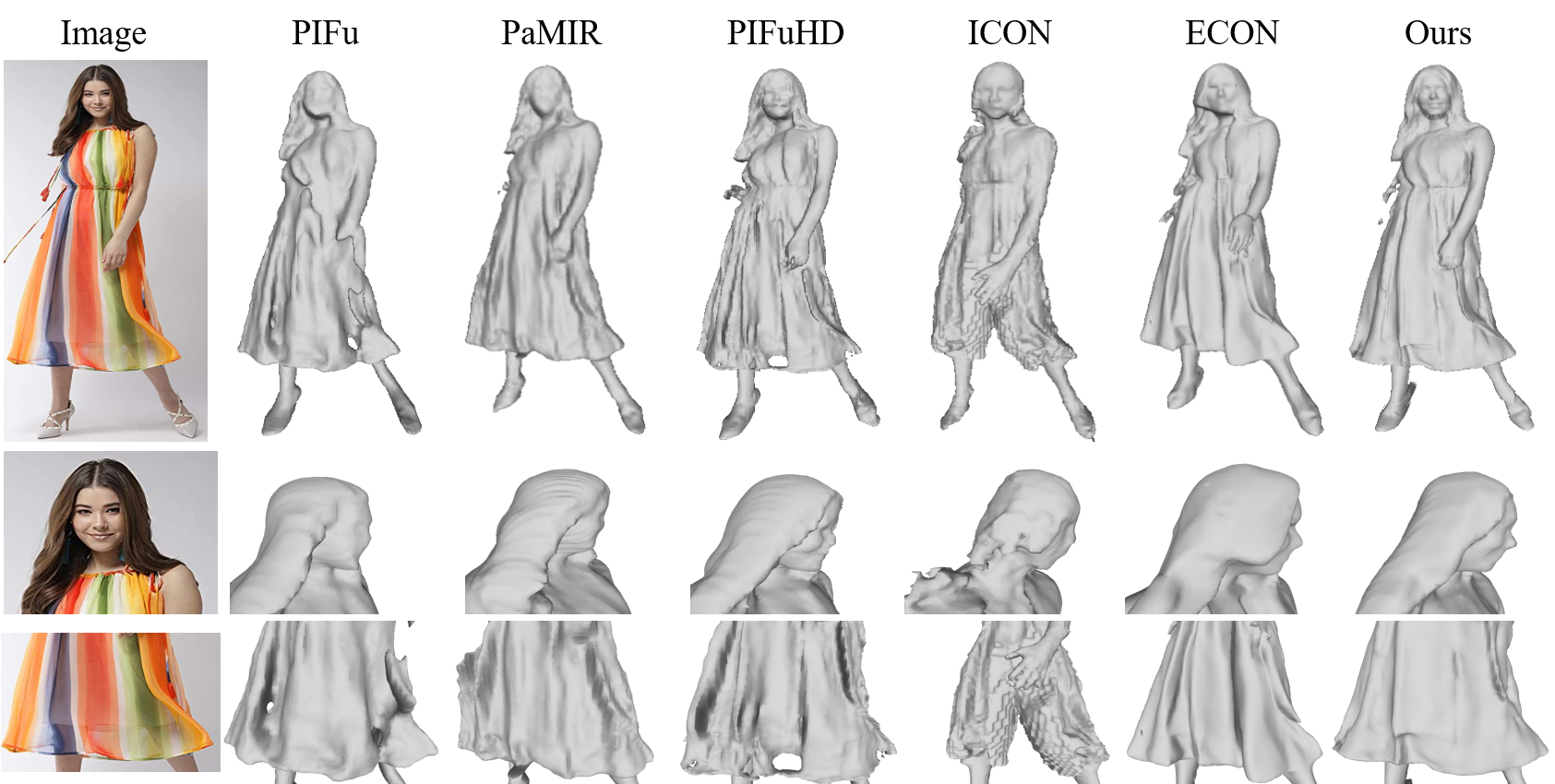}
	\caption{
	\textbf{Comparisons among different methods.} 
	PIFu takes a single color image as input (second column). 
	PIFuHD takes both image and predicted-normal maps as input (third column).
    PaMIR, ICON, and ECON all utilizes SMPL prior (from fourth to sixth line), while ECON also introduces depth maps to optimize the coarse mesh.  
    Ours takes a predicted-depth map as input (seventh column).
    The predcited-depth maps are shown in right. 
    Ours contains more completed and reasonable geometry detailed.
    }
    \vspace{-3mm}
	\label{fig_teaser}
\end{figure*}

Based on the above observations, we proposed to utilize depth estimation as our auxiliary (intermediate) input to the PIFu network, and further convert depth into effective supervisions that enable our self-supervised PIFu (SelfPIFu).
SelfPIFu consists of a normal estimator, a depth estimator and a novel self-supervised SDF-based pixel-aligned implicit function (PIFu) learning module that takes depth map as input.
The self-supervised mechanism considers two kinds of supervisions, \ie, a volume-aware and a surface-aware one, to optimize PIFu using a mixture of synthetic data with 3D human ground truth and in-the-wild images with well-estimated depth maps.
Specifically, the volume-aware self-supervision utilizes \emph{volume} supervision from point cloud data, which is transformed from the depth map.
The surface-aware self-supervision utilizes a differentiable surface renderer~\cite{liu2020dist} from SDF as supervision.
With their help, our SDF-based PIFu effectively learns convincing surface details especially for in-the-wild images.
Extensive experiments show that our method outperforms state-of-the-art human digitization approaches on both synthetic and real images.

In summary, the contributions of this work are as follows:
\begin{itemize}
  \item A novel self-supervised framework, including volume-aware and surface-aware SDF learning, is proposed to reconstruct more accurate geometry especially on real-world images.
  \item We propose to use depth map to better represent the 3D information from the image for more robust learning of implicit function, which can lead to a higher quality of human reconstruction than using normal map. 
  \item Extensive experiments and analysis on both synthetic data and real images support our claim about the advantages of depth-guided PIFu, and validate the effectiveness and superiority of the proposed framework.
\end{itemize}

The remainder of the paper is organize as follows. 
In Sec.~\ref{sec:related_work}, we briefly provide a survey of related work. 
Sec.~\ref{sec:method} first presents our depth estimator and our implicit function, then introduce the depth-guided self-supervised framework. 
Sec.~\ref{sec:experiments} conducts extensive evaluations on our SelfPIFu with SOTA. 
At last, Sec.~\ref{sec:conclusion} gives several concluding remarks.

\section{Related Work}
\label{sec:related_work}

\subsection{Singe-view Human Reconstruction}
3D human reconstruction from a single RGB image is inherently challenging due to the view occlusion and shape ambiguity. 
To address this ill-posed problem, pioneer works (\eg\ SCAPE~\cite{anguelov2005scape} and SMPL~\cite{loper2015smpl}) propose parametric models that are derived from large scanned human datasets to provide strong priors. 
However, methods based on parametric models are restricted to producing naked human bodies without garment details~\cite{anguelov2005scape,loper2015smpl,pons2015dyna,bogo2016keep,hmrKanazawa17,lassner2017unite,joo2018total,omran2018neural,pavlakos2019expressive}. 
Although \cite{zhu2019detailed} introduces a hierarchical mesh deformation network to restore detailed shapes, the results are far from the ground truth. 
Some other methods attempt to define a garment template mesh and deform it to approach the target shape of clothes~\cite{bhatnagar2019multi,tiwari20sizer,jiang2020bcnet,bertiche2020cloth3d,zhu2020deep}, but they are limited to generating specified garments and coarse details due to the fixed topology and resolution of the template. 
Besides these template-based methods, silhouette-based~\cite{natsume2019siclope} and voxel-based models~\cite{varol2018bodynet,zheng2019deephuman} are proposed for human reconstruction with arbitrary topology and geometry of clothes. 
Yet, they still produce rough geometry using a low-resolution representation due to the heavy calculation consumption.   

Recently, implicit models~\cite{chen2019learning,mescheder2019occupancy,park2019deepsdf,xu2019disn} are applied to image-based 3D reconstruction with an arbitrary resolution and fine-grained details, which dominate the field of 3D reconstruction. 
PIFu~\cite{saito2019pifu} introduces a pixel-aligned implicit function for human digitization which extracts pixel-aligned local features to recover detailed geometry. 
They further propose PIFuHD~\cite{saito2020pifuhd} with a multi-level architecture to generate higher-resolution details in line with the input image, achieving state-of-the-arts. 
However, PIFuHD requires sophisticated human models for training while the 3D dataset is extremely limited, making PIFuHD fail when applying to in-the-wild images with challenging human poses and diverse garment topology. 
In addition, PIFuHD relies on the normal generation that is not as accurate as depth estimation for 3D reconstruction.  
GEO-PIFu~\cite{geopifuhe2020} uses a structure-aware 3D U-Net to to inject global shape topology into a deep implicit function. 
Their results seem to surpass PIFu, but they still can't compare to PIFuHD as far as high frequency details are concerned. PaMIR~\cite{zheng2021pamir} brings in a parametric human model to improve the generalization ability of the implicit model, it loses some geometry details. 
ICON~\cite{xiu2022icon} utilizes SMPL-guided clothed-body normal prediction and local-feature based implicit surface reconstruction to achieve impressive results, especially in extreme poses. 
The author further propose the ECON~\cite{xiu2023econ}, which introduces intermediate depth maps based on ICON to achieve better results.
However, ICON and ECON need to perform an additional optimization process for each image.
In the same year, another work Difu~\cite{song2023difu} also introduces the depth maps as a middle variable to obtain better results.

In this paper, we utilize a robust depth generator to guide PIFu learning, which presents better generalizability than PIFuHD and achieves better performance than PaMIR, ICON \& ECON.

\subsection{Singe-view Depth Estimation}
Single-view depth estimation is a fundamental task in computer vision. Traditional methods attempt to figure out depth values from images based on hand-crafted features or geometric priors, such as stereopsis~\cite{poggio1984analysis}, camera focus~\cite{darrell1988pyramid}, shading~\cite{miller1994efficient}, and normals~\cite{nehab2005efficiently}. They tend to produce artifacts when applied to real images with various noise and varying illumination. Recent studies mainly focus on learning-based methods to obtain accurate depth estimation. Most of these works 
are about scene analysis and reconstruction~\cite{liu2015deep,laina2016deeper,wang2018learning,qi2018geonet,fei2019geo,qiu2019deeplidar,xian2020structure,yin2021learning}. 
\cite{varol2017learning} presents a large-scale synthetic dataset for human depth estimation. However, the depth is derived from the SMPL model~\cite{loper2015smpl} which misses surface details. 
\cite{tang2019neural} separates depth estimation into a smooth base shape and a residual detail shape and design a network with two branches to regress them respectively based on really scanned RGBD dataset of clothed humans. The following work~\cite{tan2020self} extends it to video-based human depth estimation by first calculating an SMPL model at each video frame (\ie\ the smooth base shape) and then learning the residual detail shape in a self-supervised manner. Even so, the details~\cite{tang2019neural,tan2020self} are not as realistic as the visualization presented in the images. 
\cite{jafarian2021learning} achieves high fidelity results by jointly learning the depth along with the surface normal and warping local geometry among video frames to enforce temporal coherence. In this paper, we also utilize a normal generation to assist our depth estimation, which can provide more accurate geometry priors for 3D human reconstruction.

\subsection{Self-supervised 3D Reconstruction}
With the limited availability of 3D data, studies are emerging to directly learn 3D shapes from input images without ground truth supervision. These methods can be roughly grouped into three categories based on differentiable rendering~\cite{kato2020differentiable}, view consistency, and frame consistency, respectively. For the single-view reconstruction, rendering-based methods propose a differentiable renderer to formulate the connection between the inferred 3D surface (\eg\ voxel, mesh, and implicit field) and its 2D renderings (\eg\ silhouette, depth, normal, and RGB images),  which can serve as image-based supervision by comparing the renderings with the input images~\cite{liu2019soft,liu2019learning,niemeyer2020differentiable,liu2020dist,jiang2020sdfdiff,hu2021self}. View-based methods attempt to apply the geometry consistency of different views to self-supervised learning for multi-view reconstruction~\cite{deng2019accurate,sanyal2019learning,wu2019mvf,shang2020self}. Similarly, Video-based methods utilize the consistency and continuity among video frames to restrict the underlying 3D shapes~\cite{mahjourian2018unsupervised,tan2020self,jafarian2021learning,wen2021self}. To improve the generalization ability of 3D human reconstruction with implicit learning, we first learn a robust depth estimation to provide accurate geometry cues and then apply the inferred depth to supervise SDF learning. We not only utilize a differentiable rendering~\cite{liu2020dist} to produce surface-aware supervision but also propose a novel volume-aware supervision based on the inferred depth, both contributing to obtaining high fidelity clothed human shapes without ground truth supervision.

\begin{figure*}[th]
\centering\includegraphics[width=0.99\linewidth]{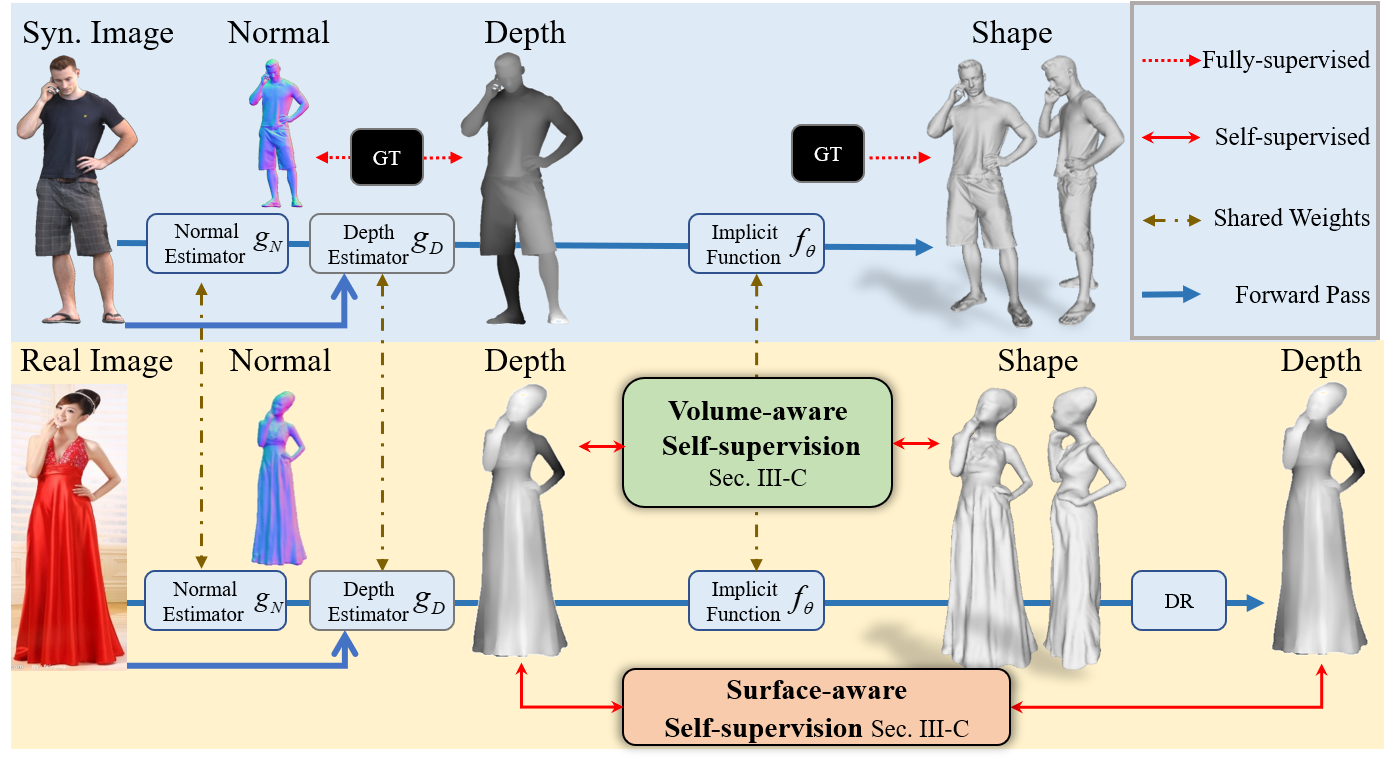}
	\caption{\textbf{Overview of our framework.}
	It consists of a normal estimator, a depth estimator, and a depth-guided SDF-based PIFu module that enables self-supervised learning. 
	For synthetic data with 3D ground truth (top row) , we use fully-supervised learning method similar to PIFuHD~\cite{saito2020pifuhd}. 
	For in-the-wild images without any labels (bottom row), we first estimate their depth maps, and then optimize the network parameters using a novel depth-guided self-supervised learning method (Sec.~\ref{sec:Self-V}).
	We visualize the image and point-cloud-like depth map in the middle of frame. 
	}
	\label{fig:pipeline}
	\vspace{-3mm}
\end{figure*}

\section{Method} \label{sec:method}

An overview of our framework is shown in Fig.~\ref{fig:pipeline}.
Given a single-view image, we first employ a normal estimator 
and a depth estimator to generate the corresponding depth and then feed the depth into a pixel-aligned implicit function (PIFu) module to predict the 3D human shape. 
For synthetic data with 3D labels, we use a fully supervised mechanism like PIFu~\cite{saito2019pifu} and PIFuHD~\cite{saito2020pifuhd} (top row).

Due to limited image-3D shape pairs as supervision, we propose a novel self-supervised training mechanism, termed as SelfPIFu, to improve the generalization ability of the existing PIFu based methods.
Specifically, the inferred depth is further used to provide \emph{volume-aware} and \emph{surface-aware} supervision through depth-guided space sampling and differentiable implicit field rendering. 
With their help, we can effectively optimize PIFu with in-the-wild images that do not have 3D geometric GT during training, resulting in largely improved generality (bottom row).

\subsection{Normal and Depth Estimation}

Depth map not only generates 3D cues as direct input for implicit learning but also indirectly provides two self-supervised mechanisms which are volume-aware SDF learning and surface-aware SDF learning for implicit training without 3D ground truth. 
To obtain accurate depth estimation, we adopt the advanced HDNet~\cite{jafarian2021learning} that jointly learns an auxiliary normal. \ie:
\begin{align}
    g_{N}(x;I) = N_{pred},  \\
    g_{D}(x;I,N) = D_{pred},
\end{align}
where $I$ represents the input image, where $x\in\mathbb{R}^2$ is the xy-location in the image, $N_{pred}$ and $D_{pred}$ represent the normal map and depth map at the corresponding location respectively. 
As we know, surface normal is responsive to the local texture, winkle, and shade~\cite{tang2019neural,wang2020normalgan}. 
Although the geometric information provided by the normal map may be incorrect, it is still useful for subsequent depth map learning. 
Like the results in \cite{tang2019neural} and \cite{jafarian2021learning}.

We train the normal and depth estimator fully supervised on RenderPeople~\cite{renerpeople} data with ground truth and minimize the following overall loss:
\begin{equation}
\begin{aligned}
L_{N} = \lambda_{cos}\cos ^{-1} \left(\left(\frac{{N_{pred}}}{\|{N_{pred}}\|}\right)   
\left(\frac{{N_{gt}}}{\|{N_{gt}}\|}\right) \right)  \\
+ \lambda_n||N_{gt}-N_{pred}||_1
\label{eq:ln}
\end{aligned}
\end{equation}

\begin{equation}
L_{D} = ||D_{gt}-D_{pred}||_1,
\end{equation}
where $\lambda_{cos}$ and $\lambda_n$ relative weights between losses, $N_{pred}$ and $D_{pred}$ are predicted normal map and depth map, $N_{gt}$ and $D_{gt}$ represent the ground truth. The depth estimator achieves strong generalization on in-the-wild images, just like Fig.~\ref{fig:intro_depth2pc}, more results will be shown in the next Section.

\subsection{SDF-based Pixel-aligned Implicit Function from Depth}

We use a signed distance function (SDF) instead of occupancy to represent 3D geometry, because SDF field is continuous in the 3D space and can interpolation reasonable details when using marching cube to get a mesh.
After we obtain a robust depth estimator to infer geometry information from the input image, the goal of 3D human digitization is to model a function $f$ which can calculate the SDF value $s$ of an arbitrary query point $p\in\mathbb{R}^3$ under the observation of an inferred depth map, \ie:
\begin{align}
    f_\theta(g(x,D_{pred}), z(p)) = s, s\in\mathbb{R},
\end{align}
where $x=\pi(p)$ is the 2D projection of query point $p$, $g(x,D)$ is depth feature at $x$, $z(p)$ is the depth value of $p$ in the camera coordinate space, and $\theta$ is the trainable parameters of our PIFu module.

Our implicit function module $f_\theta$ consists of a depth encoder (instead of an image encoder) 
and an SDF decoder. 
The depth encoder is a fully convolutional network using an hourglass architecture~\cite{newell2016stacked}, while the SDF decoder is made up of multi-layer perceptrons (MLPs). 
Implementation details can be found in Sec.~\ref{sec:experiments}.

\begin{figure*}[t]
	\centering
	\includegraphics[width=0.95\linewidth]{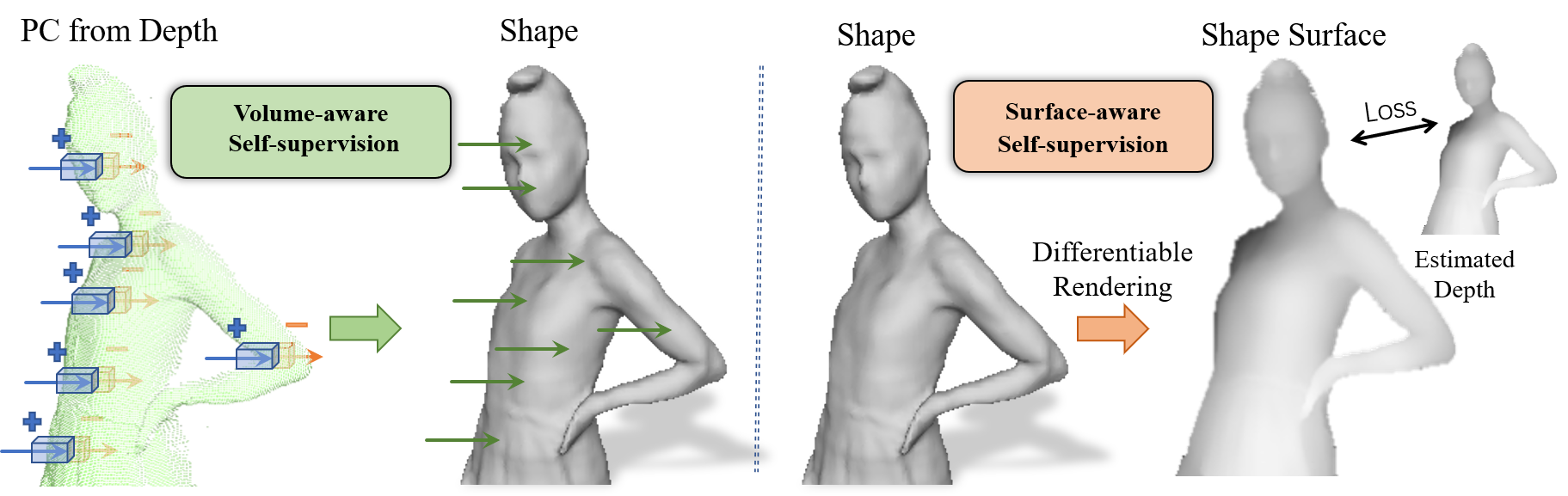}
	\caption{\textbf{Depth-guided self-supervised learning.} 
	\emph{Volume-aware} self-supervised SDF learning (left) utilizes on-/near-surface points converted from the estimated depth map to provide pseudo \emph{volume} supervision.
	\emph{Surface-aware} self-supervised SDF learning (right) imposes a \emph{surface}-wise self-supervision by comparing DIST~\cite{liu2020dist} rendered and estimated depth maps.
	More details are described in Sec.~\ref{sec:Self-V}.
	}

	\label{fig:self_sup}
 	\vspace{-3mm}
\end{figure*}

\subsection{Depth-guided Self-supervised Learning} \label{Self-learn}

The training of PIFu requires plenty of 3D clothed human shapes spanning diverse identities, poses, and complex garment geometry. 
However, 3D data is limited and expensive to acquire. 
On the other hand, sufficiently diverse social media images are available online and the state-of-the-art depth estimator usually gives accurate enough depth estimates with convincing geometry details.
So we propose two self-supervised mechanisms in Fig.~\ref{fig:self_sup} to make best use of the well-estimated depth maps (e.g. Fig.~\ref{fig:compare_dn_details}) to improve the accuracy, robustness, and generalization ability of the SDF-based PIFu. 

\vspace{2mm}
\noindent\textbf{Volume-aware self-supervised SDF learning.} \label{sec:Self-V}
Inspired by recently neural implicit shape modeling works such as IGR~\cite{IGR}, we propose to utilize \emph{volume supervision} provided by the point cloud, which is transformed from the estimated depth map, as pseudo labels to better optimize/regularize the implicit field.
Specifically, we back-project the depth map into the camera coordinate space using orthogonal camera model to obtain 3D points. 
Because these points should be on the underlying human surface, their SDF values are set to 0. In addition, the variation of SDF value near the surface is supposed to be stable due to the continuity of the SDF field. 
We can randomly sample $N$ points along the camera view direction with a threshold $\sigma$ to generate pseudo labels for training. Specifically, we assign points far from the viewpoint with negative values and points close to the viewpoint with positive values. 
The absolute values are equal to the distances from the sampled points to the surface.

After obtaining the pseudo labels, we utilize them to supervise our PIFu learning with a $L_1$ loss. Note that SDF is good at preserving geometric details but its learning is a classic regression task that is harder to train than a classification one, e.g., the occupancy learning. 
Therefore, We increase the penalty for results with opposite signs.
Our volume-aware loss function $L_{vol}$ can be defined as

\begin{equation}
\begin{aligned}
    L_{vol} = \sum_{i}^{N}||s_i - s_i^{gt}||_1 + 
    \lambda_m(\sum_{i}^{N}||s_i - s_i^{gt}||_1 \times M_i) 
\end{aligned}
\end{equation}

\begin{equation}
M_i=\left\{\begin{aligned}
1, & \text { if } s_i \times s_i^{gt}<0 \\
0, & \text { if } s_i \times s_i^{gt} \geq 0
\end{aligned}\right.
\end{equation}

where $s_i$ is the SDF value of the sample point $p_i$, $M_i$ is the opposite signs penalty mask, $N$ is the number of sample points, and the superscript ``$gt$'' means the ground truth, \ie\ the pseudo labels. 

\noindent\textbf{Surface-aware self-supervised SDF learning.}
In addition to the \emph{volume} supervision from discrete samples, we also propose continuous 
\emph{surface} supervision based on an differentiable renderer DIST~\cite{liu2020dist}. 
DIST uses a differentiable sphere tracing algorithm to render the underlying surface of an SDF field to 2D observations including a depth map. 
We modify DIST, which originally uses a perspective projection, to be compatible with our SelfPIFu, which uses orthogonal projection\footnote{We use parallel ray tracing instead of sphere tracing.}
We measure the difference between the DIST rendered depth $D_r$ and the estimated depth $D_{pred}$ (from the depth estimator $g_{D}$) with a $L_2$ loss function:
\begin{align}
    L_{surf} = ||D_{r}-D_{pred}||_2.
\end{align}

In our experiment, we train the PIFu module with a weighted loss $L$, \ie\
\begin{align}
    L = L_{vol} + \lambda L_{surf},
\end{align}
where $\lambda$ is a weight of the loss $L_{surf}$. 
Both volume-aware and surface-aware self-supervised SDF learning mechanisms contribute to the self-supervision of single-view 3D human reconstruction (see Table~\ref{tab:comp_input} and Fig.~\ref{fig:ablation}).

\noindent\textbf{Difference with PIFuHD.}
In contrast to PIFuHD, our PIFu module presents three main differences: 
(i) Different input. It takes the inferred depth (not normal) as input since image-based depth estimation can recover more correct geometry information than the normal generation. 
(ii) Different representation. It adopts SDF to represent a 3D shape since SDF is consecutive and can capture more detailed geometry than an occupancy field. 
(iii) Different architecture. Due to the contribution of depth and SDF, it uses only one-level PIFu and a shallow network to achieve comparable details with PIFuHD.

\section{Experiments}
\label{sec:experiments}

We conduct extensive experiments on both synthetic and real data to compare SelfPIFu with other state-of-the-art (SOTA) methods. 
Quantitative and qualitative results have justified the effectiveness of SelfPIFu and its individual components.

\subsection{Datasets, Metrics, and Implementation Details}

\noindent\textbf{Datasets.} 
We use two types of data in our experiments: 
1) synthetic clothed human data from RenderPeople with rendered images and their corresponding 3D shape, and 2) real-world images or videos without 3D ground truths. 
On the synthetic dataset, we use 278 subjects of RenderPeople data for training and 22 subjects for evaluation. 
Similar to the process in PIFuHD~\cite{saito2020pifuhd}, for each training subject, we render an image, a depth map, and a normal map every azimuth degree with varying lighting , producing 360 training triplets in total. 
The depth maps, which are not utilized in PIFu/PIFuHD, are rendered using OpenGL~\cite{opengl}.
On the realistic dataset, we use a total of 15,320 real images, of which 1,400 are collected from the TikTok dataset~\cite{jafarian2021learning}, and 13,920 are crawled from the Internet by ourselves to include more diverse identities, poses, and appearances/clothes, among which 320 images are randomly chosen for evaluation. 

\vspace{2mm}
\noindent\textbf{Evaluation Metrics.}
To evaluate the quality of the generated 3D human shape, we adopt commonly used Chamfer distance (CD), point-to-surface distance (P2S), and intersection-over-union (IoU) between the generated shape and the ground truth. 
For real images without 3D ground truth, we visualize the generated 
results.

\begin{figure}[h!]
	\centering
	\includegraphics[width=0.95\linewidth]{ 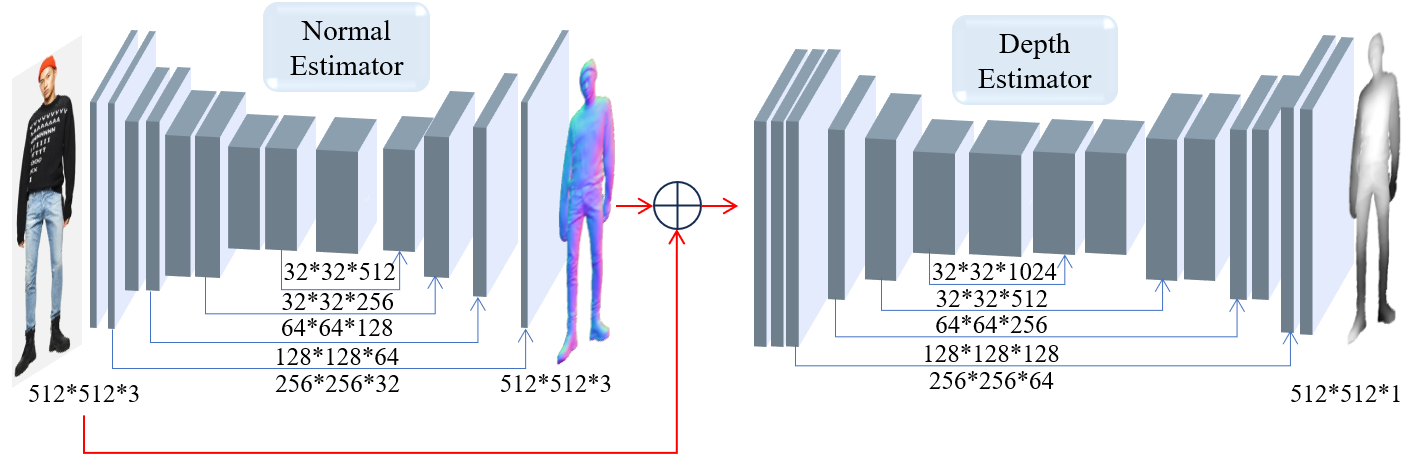}
	\caption{The structures of normal and depth estimator. All kernel size of convolutions set to 3*3. 
	}
 \vspace{-3mm}
	\label{fig:model_nd}
\end{figure}

\noindent\textbf{Implementation Details.}
The design of the normal and depth estimator in SelfPIFu is shown in Fig.~\ref{fig:model_nd}.
In the implicit function $f_\theta$ in Fig.~\ref{fig:pipeline}, 
the encoder adopts hourglass \cite{newell2016stacked} and uses group normalization \cite{wu2018group} instead of batch normalization \cite{jackson20183d} as in PIFuHD \cite{saito2020pifuhd}. 
The feature maps output by the encoder is 128$\times$128$\times$256, where 256 is the channel dimension, 
and is fed into an MLPs with \{257, 512, 256, 128, 1\} neuron(s) each layer with skip connections at the second and third layers. 

For the training on the synthetic data, we train the network for 100 epochs using an Adam optimizer with a learning rate (LR) of 0.0001 and batch size of 8. 
%
After the whole frame work is pretrained, we freeze the normal and depth estimator, and finetune the implicit function $f_\theta$ using the proposed self-supervised training mechanism on both the synthetic data and real world images.
In this stage, the LR is reduced to one-tenth and the batch size is set to 2. 
All images used for training are resized to 512$\times$512 as in PIFuHD. 
In the implicit function training, we sample 8000 points with a mixture of uniform sampling and importance sampling around the surface with standard deviations of 3cm. 
In the self-supervised training, $\lambda$ is set to 0.618. $\lambda_{cos}$ and $\lambda_{n}$ are set to 1.25 and 1 in \eqref{eq:ln}. 
The threshold $\sigma$ in \ref{sec:Self-V} is set to 1.5cm.  

We follow the \cite{jafarian2021learning} to train the depth estimator. It is worth noting that during the training process of the depth estimator, incorporating a normal map in the network's input and adding constraints from the surface normal map to the depth map can effectively improve the accuracy of depth map estimation.



\subsection{Evaluations} \label{sec:eval}
\noindent\textbf{Depth v.s. Normal as Input.} 
We first experimentally validate our claim that the depth map is a better choice as input for human reconstruction. 

\textbf{\textit{Comparison on Synthetic data}}: Two groups of quantitative experiments are performed on the RenderPeople dataset to show the superiority of using depth as input over using normal as input. 
Notice that in order to avoid the influence of other factors, such as normal estimator and depth estimator, in \textit{Depth v.s. Normal as Input} part, all inputs are ground-truth data rendered from textured meshes. 

In the first group, we test the usefulness of various input combinations, including RGB image (denoted as ``I''), normal map (denoted as ``N''), depth map (denoted as ``D''), and their combinations (e.g. ``ID'' for using image and depth as input).
We follow the standard fully-supervised training process of PIFu \cite{saito2019pifu} (i.e. w/o the proposed self-supervised training branches in Sec.~\ref{Self-learn}) in these experiments.
The reconstructed shape of each variant is evaluated by the IoU with the ground truth. 
During the training process, we randomly sample 10k near-surface points from the GT mesh of the synthetic data, and computed the IoU on the training and testing sets, respectively.

\begin{figure}[htb]
	\centering
	\includegraphics[width=0.99\linewidth]{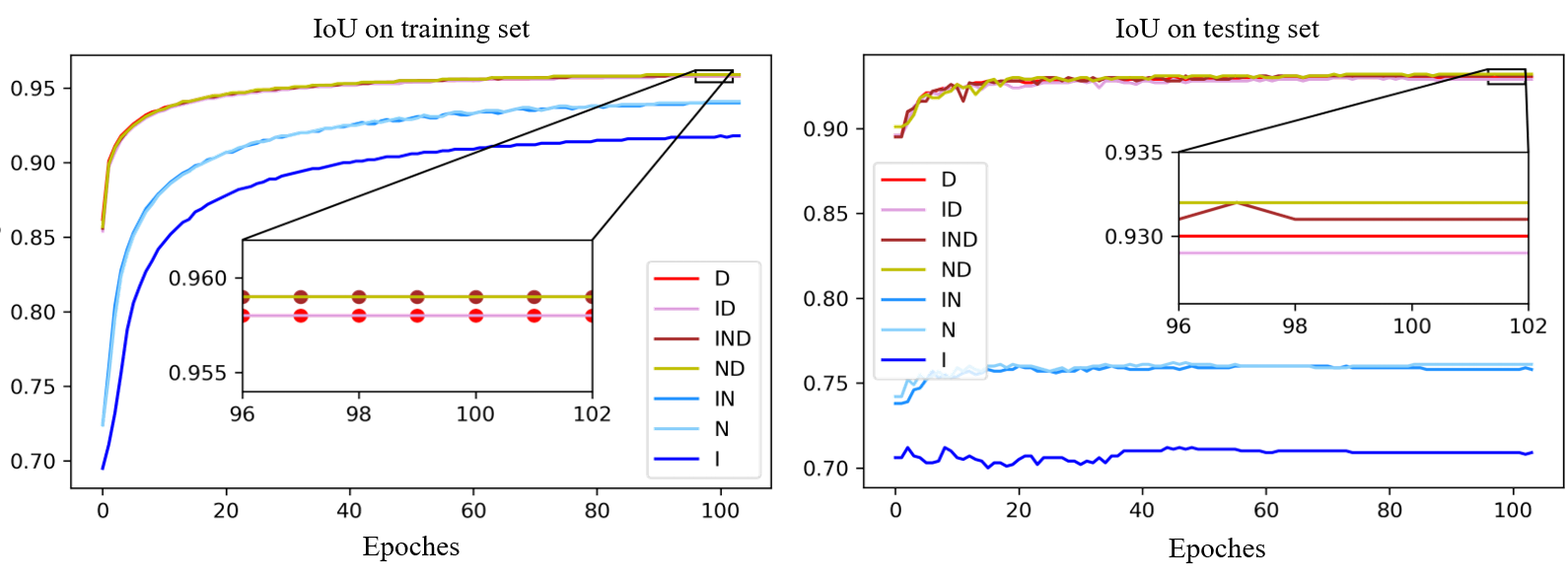}
	\caption{Comparison of using different input information as input, including image (denoted as ``I''), normal map (``N''), depth map (``D''), and their combinations (e.g. ``ID'' for image and depth). 
	Training (top) and test (bottom) loss curves during training are plotted.
	}
	\vspace{-2mm}
	\label{fig:IOU_T}
\end{figure}

The reconstruction results measured with IoU are shown in Fig.~\ref{fig:IOU_T}. 
In training set, using only image (i.e. ``I'') as input results in 91.8\% IoU.
Including normal as input (``IN'' or ``N''), the accuracy increases to around 94.0\%.
Including depth as input (``D'', ``ND'', ``IND''), the accuracy increases to around 95.8\%.
Clearly, including depth as input significantly improve the reconstruction IoU. 
One possible reason is that the gap between depth and 3D geometry is smaller than the gap between normal/image and 3D geometry, which means depth is more informative than normal and normal is more informative than image for this shape reconstruction.
When an more informative exists in the input, the extra less informative input modality can only provide minimal gain (e.g. 95.8\% for ``D'' vs 95.8\% for ``ID'', 95.9\% for ``ND'' vs 95.9\% for ``IND'')

\begin{figure*}[t]
	\centering
	\includegraphics[width=0.99\linewidth]{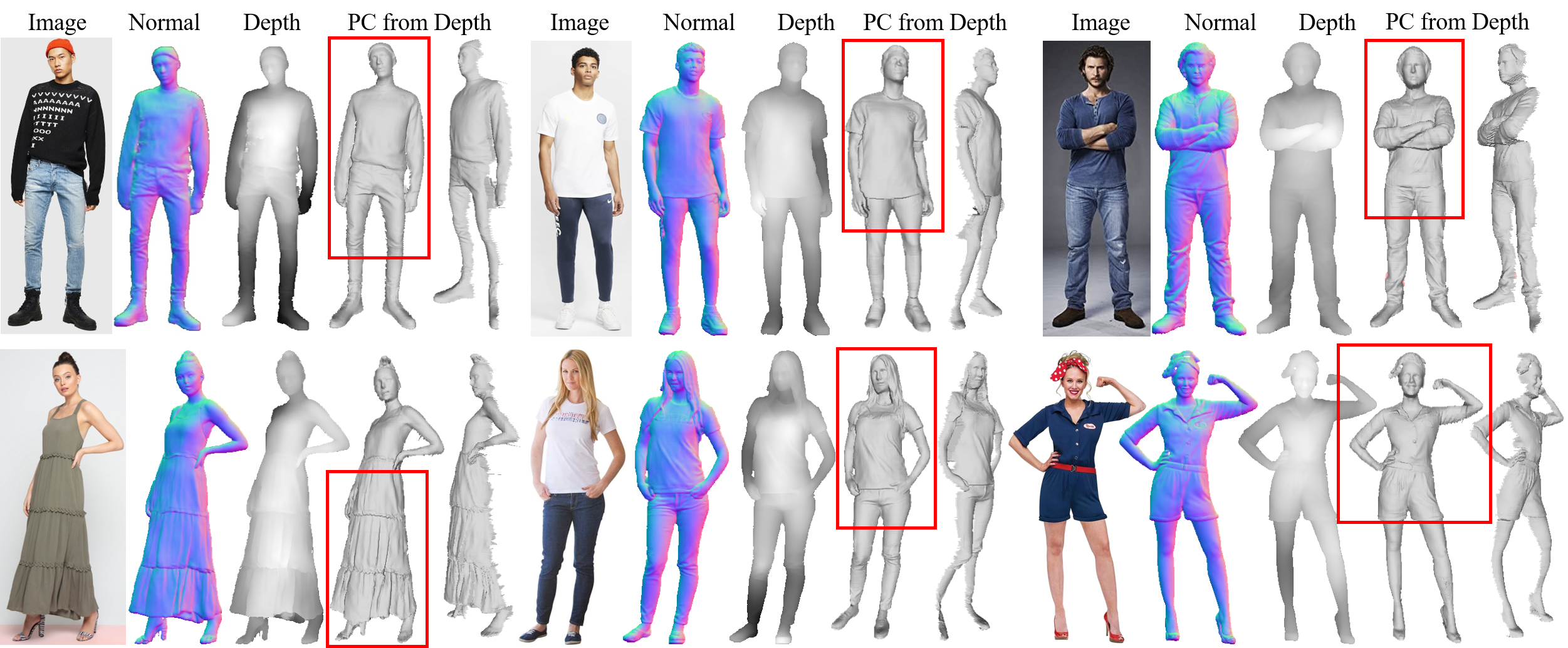}
    \caption{Visual comparison of the estimated normal maps and depth maps on real world images. 
	Even when tested on in-the-wild images, the estimated depth maps and point clouds (PC) contain high-fidelity details, which are effectively transferred into the final 3D geometry by the proposed depth-guided self-supervised learning module.
	See Sec.~\ref{Self-learn} for details and Fig.~\ref{fig:vis_NvsD} for comparisons. 
	}
 \vspace{-3mm}
	\label{fig:compare_dn_details}
\end{figure*}

Compare with training set, we care more about the results on testing set. 
Similar trend has been observed on the test set while the gap between depth and other modality becomes even bigger, demonstrating that including depth as input could substantially improve the generalization ability.
For example, the IoU is above 93.0\% when including depth as input. While the IoU drastically drops under 76\% when removing depth as input.
For example,when the input is a single image, the reconstruction IoU is only around 70.8\%. 
With a normal map as or among the input, the IoU rises to around 75.8\%.

\begin{figure}
	\centering
	\includegraphics[width=0.95\linewidth]{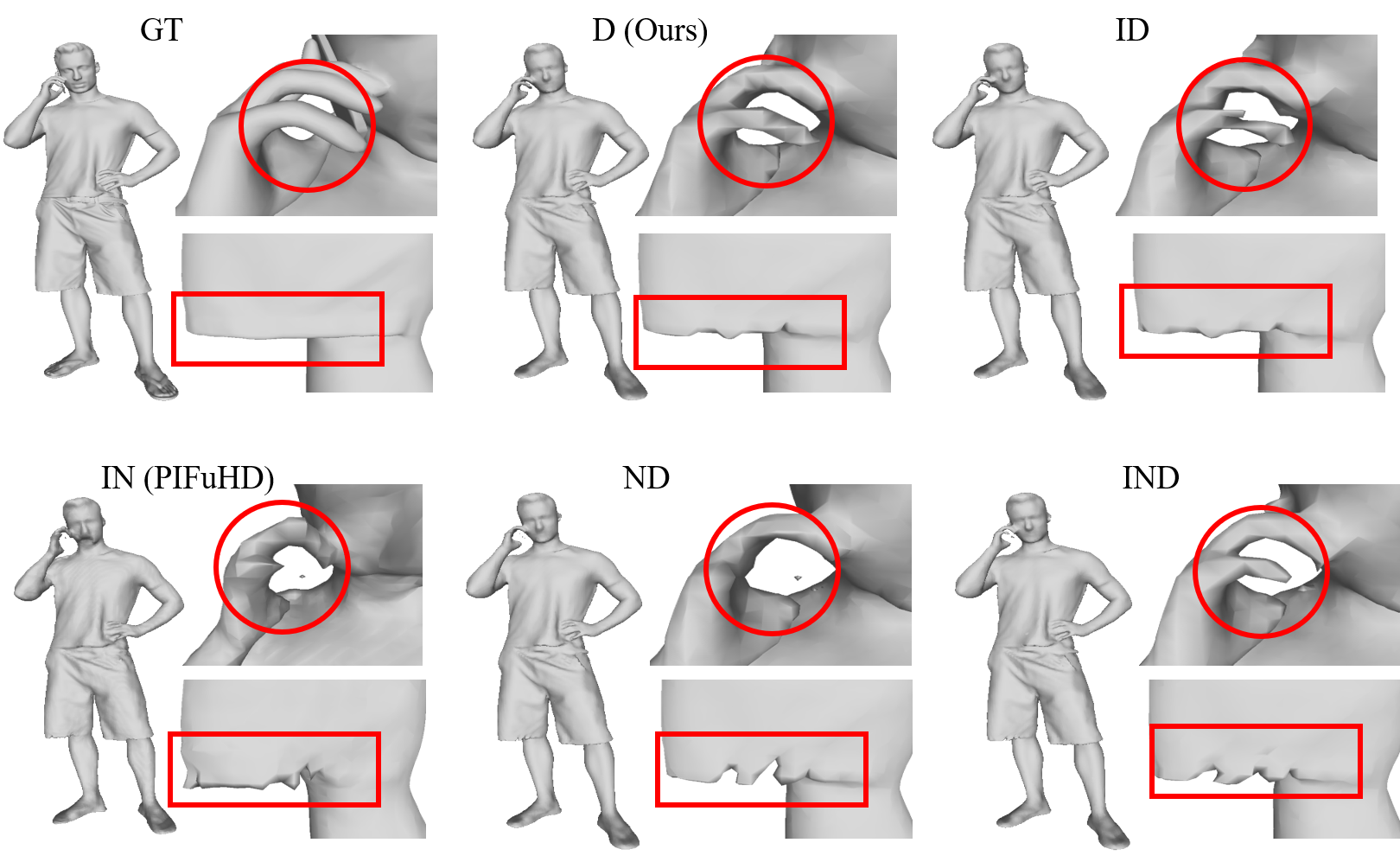}
	\caption{
	Comparison of using different input information as input on synthetic data with GT mesh, including depth map (denoted as ``D'') and the combinations of  image, depth map, and normal.  (e.g. ``ID'' for image and depth)
	}
 \vspace{-3mm}
	\label{fig:choose_D_image}
\end{figure}

Visual comparison of different kinds of input combinations on synthetic data in shown in Fig.~\ref{fig:choose_D_image}.
Using the depth map as input, the shape is the most complete and visually closer to GT.
In particular, we observe that in the edge region, the shape becomes incomplete when the input contains a normal or an image. 
This may be because the depth provides sufficient information about the frontal space geometry. 
Additional image and/or normal input instead introduce additional uncertainties that affect the prediction of the geometry by the implicit field function.
Considering the visualization results of the reconstruction, we use the depth map as the only input to our SlefPIFu.

In the second group, we follow the pipeline of SelfPIFu, where the proposed depth-guided self-supervision is included to make better use of the depth information for reconstruction. 
This module further increases IoU up to 0.5\% (e.g. 93.0\% to 93.5\% for ``D'')
The reconstruction IoUs of using different types of intermediate are listed in the last column Table~\ref{tab:comp_input}. 
These results show the great potential of using depth for high-quality reconstruction, not only in a supervised training scenario, but also when using self-supervision methods for further improvements. 

\begin{table}
    \centering
    \begin{threeparttable}
    \caption{The IoUs (\%) between reconstructed shapes and the ground truths on the RenderPeople dataset with respect to different input combinations.}
    {\def\arraystretch{1} \tabcolsep=1.2em 
        \begin{tabular}{ccc|c|cc}
        \toprule
        Image & Normal & Depth & & IoU (\%) & $\text{IoU}_\text{self}$ (\%)  \\
        \midrule
        \checkmark &  &  & I &70.8 & ~\\
         & \checkmark &  & N &76.1 & ~\\
         &  & \checkmark & D &93.0 & 93.5~(+0.5)\\
        \checkmark & \checkmark &  & IN & 75.8 & ~\\
        \checkmark &  & \checkmark & ID & 92.9 & 93.5~(+0.5)\\
         & \checkmark & \checkmark & ND & 93.1 & 93.5~(+0.4)\\
        \checkmark & \checkmark & \checkmark & IND & 93.2 & 93.5~(+0.3)\\
        \bottomrule
        \end{tabular}
    }
    \begin{tablenotes}
    \scriptsize
        \item We compare using different input information, including image (denoted as ``I''), normal map (``N''), depth map (``D''), and their combinations (e.g. ``ID'' for image and depth). ``$\text{IoU}_\text{self}$'' means adding our proposed depth-guided self-supervised module during training.
    \end{tablenotes}
    \vspace{-3mm}
    \end{threeparttable}
    \label{tab:comp_input}
\end{table}

\noindent\textbf{Depth Map Contains Details.} 
To better demonstrate the details contained in the estimated depth map, we convert depth mesh to 3D point cloud and render it from different views.
Specifically, we first project depth to 3D (with orthogonal camera model) to get the 3D point cloud, and then generate a mesh from it for rendering.
Fig.~\ref{fig:compare_dn_details} shows the estimated normal maps and depth maps from in-the-wild images, and rendered point cloud image from depth.


\begin{figure*}[ht!]
	\centering
	\includegraphics[width=0.99\linewidth]{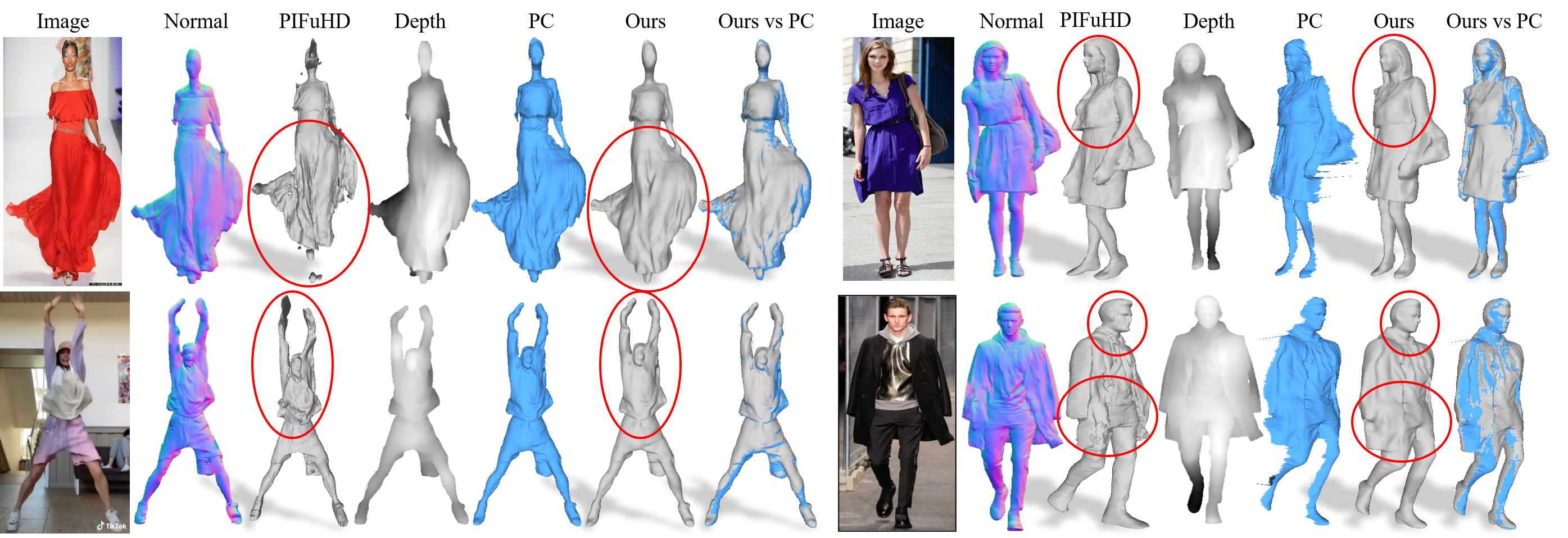}
	\caption{
	Visualization of reconstruction results on PIFuHD and Ours.
	PIFuHD sometimes generates false disconnected regions when applied on real images (first row). 
	In comparison, Ours can recover a complete human shape with almost no small shards.
	Furthermore, PIFuHD usually reconstructs false details, {\eg} unrealistic face features and clothes wrinkles (second row).
	}
    \vspace{-3mm}
	\label{fig:vis_NvsD}
\end{figure*}

\begin{figure}[t]
	\centering
	\includegraphics[width=0.98\linewidth]{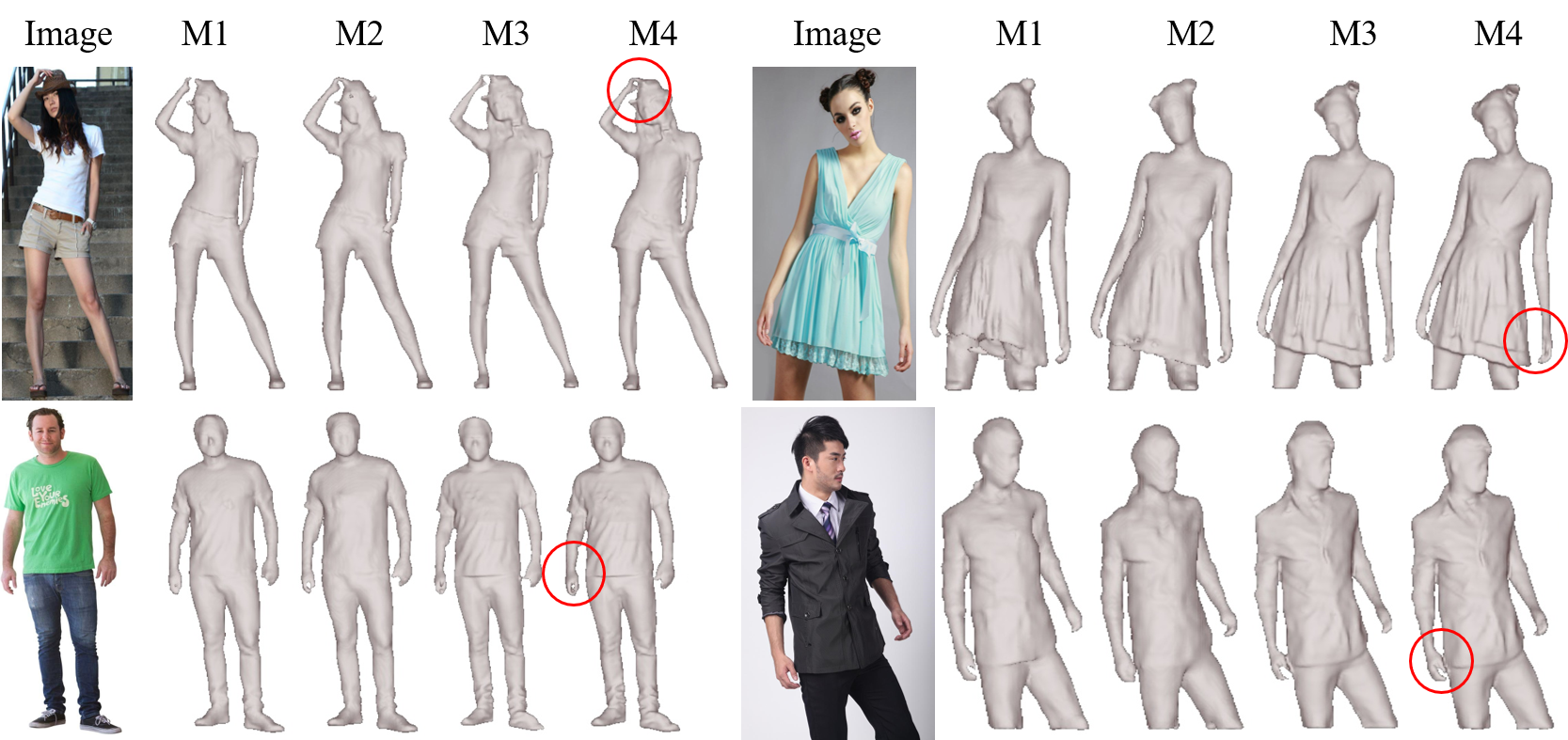}
	\caption{\textbf{Ablative study on the self-supervised mechanism.} 
	``M1-M4'' represent using normal map as input (M1), using depth map as input (M2), using depth map as input plus surface-aware self-supervision (M3), and using depth map as input plus both surface-aware and volume-aware self-supervision (M4).}
	\label{fig:ablation}
 \vspace{-3mm}
\end{figure}

\begin{table}
    \centering
    \caption{User Study on different variants of our SelfPIFu. }
    \begin{threeparttable}
    {\def\arraystretch{1} \tabcolsep=1.4em 
        \begin{tabular}{l|llll|c}
        \toprule
    	& N & D & S & V & Score    \\
    	\midrule
    	M1 & \checkmark & & &         &      0.07     \\
    	M2 & & \checkmark & &        &       0.08     \\
    	M3 & & \checkmark & \checkmark &          &       0.23     \\
    	M4 & & \checkmark & \checkmark & \checkmark &      \textbf{0.63}    \\ 
        \bottomrule
        \end{tabular}
    }
    \label{tab:use_abl}
    \begin{tablenotes}
    \scriptsize
        \item ``N'', ``D'', ``S'', ``V'' denote normal, depth, surface-aware, and volume-aware self-supervision, respectively.
    \end{tablenotes}
    \end{threeparttable}
    \vspace{-3mm}
\end{table}

\textbf{\textit{Comparison with PIFuHD on Real World Images}}: We visualize the comparison results of PIFuHD (normal-based) and Ours (depth-based) in Fig.~\ref{fig:vis_NvsD}.
PIFuHD sometimes generates false disconnected regions when applied on real images (first row in Fig.~\ref{fig:vis_NvsD}). 
In comparison, Ours can recover a complete human shape with almost no small shards. 
Furthermore, PIFuHD usually reconstructs false details, {\eg} unrealistic face features and clothes wrinkles (second row in Fig.~\ref{fig:vis_NvsD}).
In comparison, PIFu-D can better handle such cases with complete shapes and more plausible details.

\begin{figure*}[htb]
	\centering
	\includegraphics[width=0.99\linewidth]{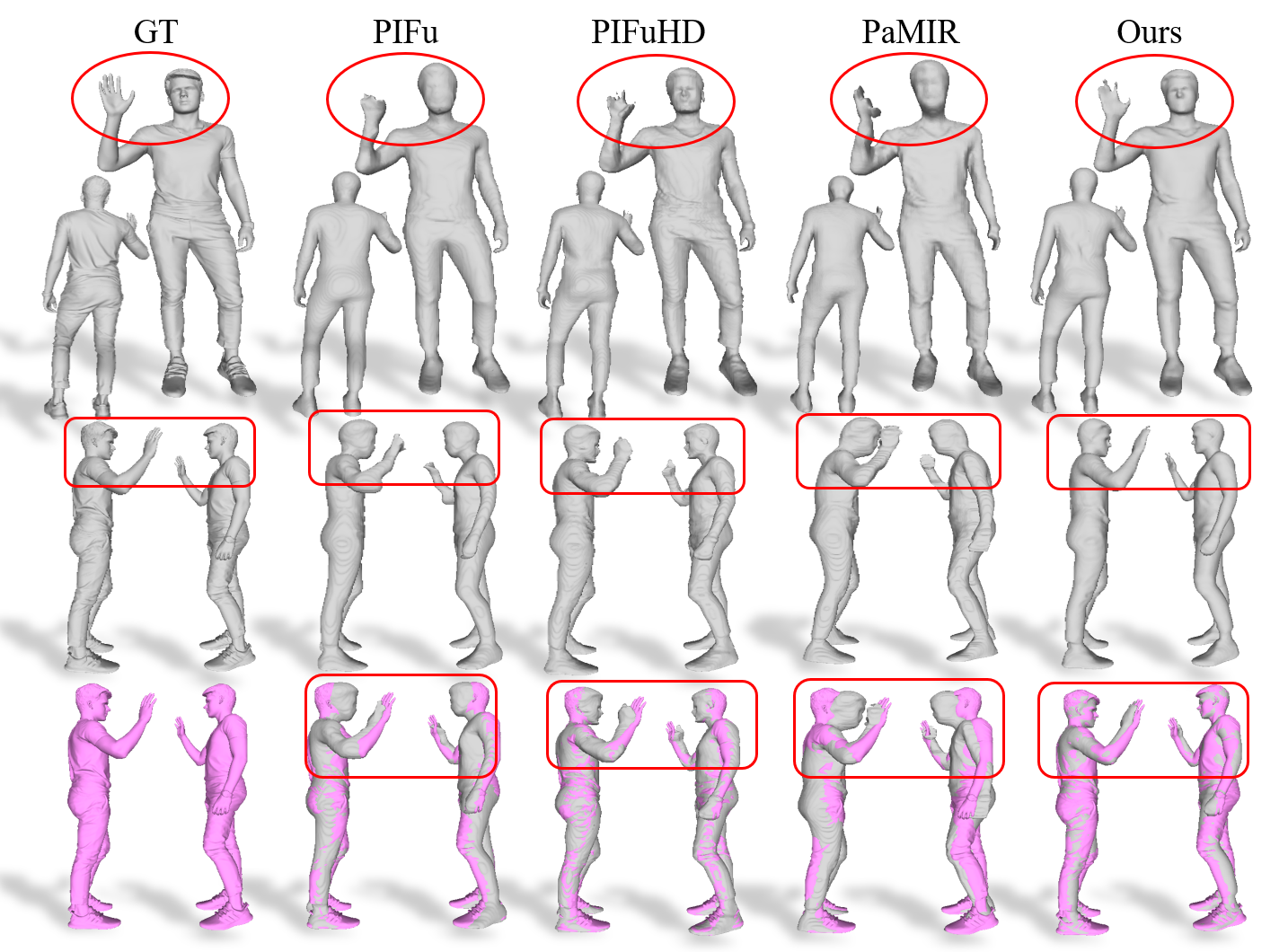}
	\caption{\textbf{Visual comparison on synthetic data.}
	We compare reconstructed meshes from PIFu~\cite{saito2019pifu}, PIFuHD~\cite{saito2020pifuhd}, PaMIR~\cite{zheng2021pamir},
    ICON~\cite{xiu2022icon}, 
    ECON~\cite{xiu2023econ},and ours. 
	The first row shows front- and back-view results. 
    The second row shows two side-view results and overlaid on GT to demonstrate the reconstruction errors.
    }   
    \vspace{-3mm}
	\label{fig:compare_syn}
\end{figure*}

\begin{figure*}[htb]
	\centering
	\includegraphics[width=0.9\linewidth]{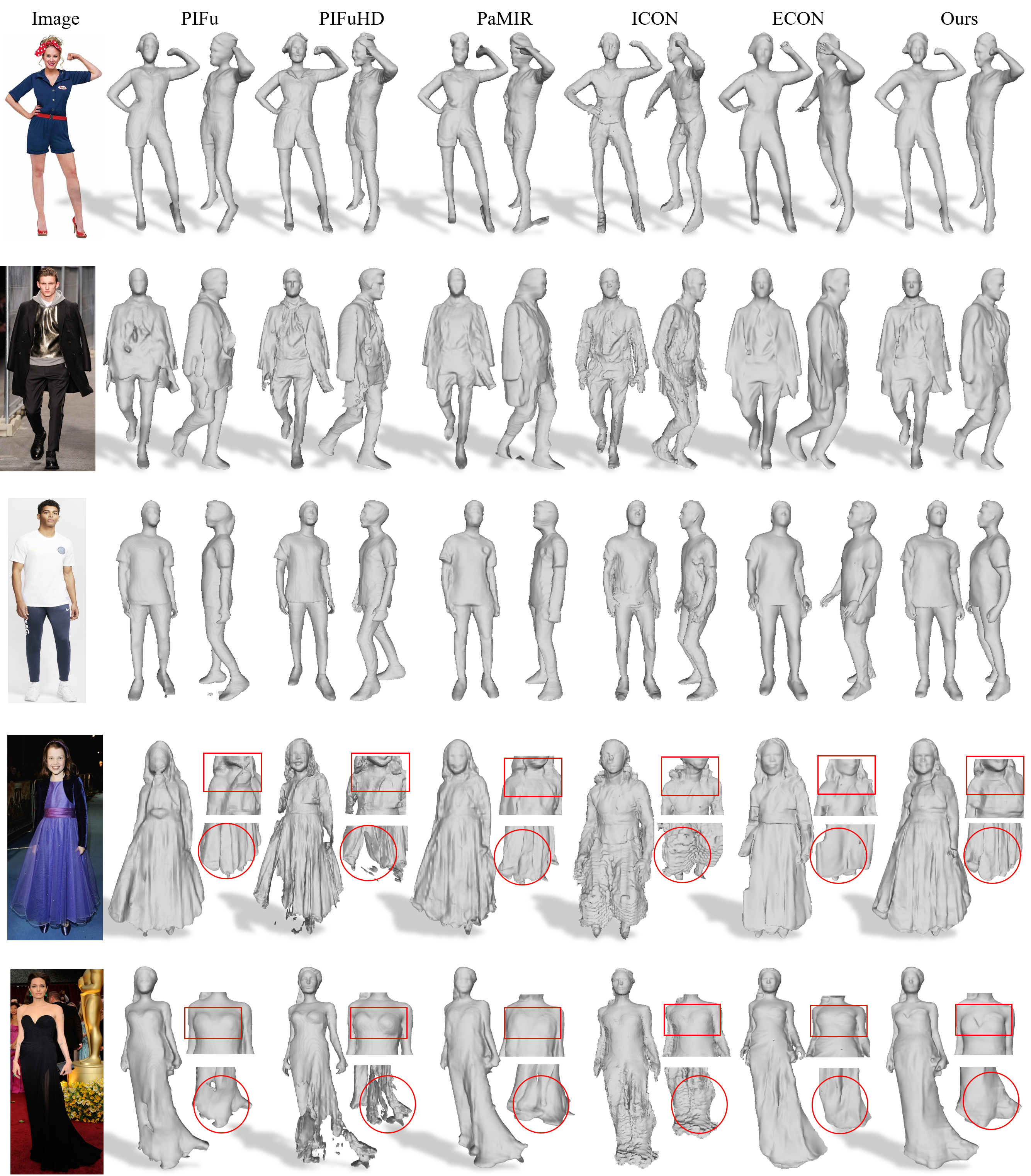}
	\caption{\textbf{Visual results on real world images.} Reconstructions from PIFu~\cite{saito2019pifu}, 
	PIFuHD~\cite{saito2020pifuhd}, 
	PaMIR~\cite{zheng2021pamir}, 
    ICON~\cite{xiu2022icon}, 
    ECON~\cite{xiu2023econ}, and ours are shown. 
 	Zoom in to see more details. 
	}
	\label{fig:compare_real}
 \vspace{-3mm}
\end{figure*}

\vspace{2mm}
\noindent\textbf{Ablation Study on the Self-supervision Mechanism.} 
The proposed self-supervision mechanism contais two parts, i.e. surface-aware and volume-aware SDF learning. 
Ablative experiments are conducted on synthetic data to demonstrate the effectiveness of either of them, respectively. 
We visualize the reconstructed meshes of several samples that are randomly chosen from the collected real images in Fig.~\ref{fig:ablation}, where four results are obtained by taking as input 
normal map only (denoted as M1), depth map only (M2), depth map plus surface-aware self-supervision (M3), and depth map plus both surface-aware and volume-aware self-supervision (M4), respectively. 
It is shown that both two self-supervision mechanisms can improve the reconstruction quality when using the depth map as the intermediate input, 
especially in the details of reconstruction. 
We also conduct a user study for human evaluation on the reconstruction results. 
10 samples are randomly chosen and evaluated by more than 50 persons for evaluation. 
The participants are asked to choose the best one among four reconstruction results (shuffled differently for each set) given the original single-view human image. 
The ratio of votes for each version is computed as its final evaluation score. 
The results of this user study are listed in Table~\ref{tab:use_abl}, which show the improvements brought by the proposed self-supervision mechanism. 


\subsection{Comparison with the State-of-the-art}

\noindent\textbf{Reconstruction of Synthetic Images.} 
We implement five SOTA methods, PIFu~\cite{saito2019pifu}, PIFuHD~\cite{saito2020pifuhd},  PaMIR~\cite{zheng2021pamir}, ICON~\cite{xiu2022icon}, ECON~\cite{xiu2023econ}, for comparison with the proposed SelfPIFu. 
On the RenderPeople dataset, the evaluation results on three metrics are presented in Table~\ref{tab:syn}. 
As seen, our SelfPIFu outperforms all listed methods consistently on all three metrics, achieving a new SOTA performance on RenderPeople with an average of 0.7934cm for Chamfer distance, 1.5377cm for P2S distance, and 89.03\% for IoU. 
It surpasses PIFuHD by around 0.4cm, 0.54cm, and 20.08\% on the Chamfer distance, P2S distance, and IoU, respectively. 

\begin{table} 
    \centering
    \caption{Quantitative evaluation on the Renderpeople dataset. 
    }
    {\def\arraystretch{1} \tabcolsep=0.9em 
        \begin{tabular}{l|ccc}
        \toprule
    	 Method & CD (cm) &  P2S (cm)  & IoU (\%) \\
    	\midrule
    	PIFu~\cite{saito2019pifu}     &  1.2473  &  2.4219  &  72.24  \\
    	PIFuHD~\cite{saito2020pifuhd} &  1.1976  &  2.3435  &  68.23   \\
    	PaMIR~\cite{zheng2021pamir}   &  2.2563  &  4.4823  &  65.09   \\
        ICON~\cite{xiu2022icon}   &  1.7857  &  3.4812  &   56.77  \\
        ECON~\cite{xiu2023econ}   &  2.0607  &  4.4395 &  60.42   \\
    	\textbf{SelfPIFu} &  \textbf{0.7934}  &  \textbf{1.5377}  &  \textbf{89.03} \\ 
        \bottomrule
        \end{tabular}
    }
    \label{tab:syn}
\end{table}

In addition to objective comparisons, we also made visual comparisons in Fig.~\ref{fig:compare_syn}.
In frontal view, the differences between the results of the different methods are small, which seems to be different from the results of the objective comparison.
However, their differences are very obvious from the side-view.
The results of PaMIR, ICON and ECON show significant offset, like head, hand and foot due to depth ambiguity, which is effectively addressed by our SelfPIFu since it uses depth map as input and, more importantly, is trained on more diverse in-the-wild images enabled by the proposed depth-guided self-supervised learning.

\noindent\textbf{Reconstruction of Real Images.} To compare with SOTA methods on the human reconstruction of real images, we provide qualitative results and also conduct user studies for human evaluation.

\textbf{\textit{Qualitative Comparison}}: 
we visualize the reconstruction result of SelfPIFu and other SOTA approaches.
As shown in Fig.~\ref{fig:compare_real}, the reconstruction products from SelfPIFu are of higher quality than others. 
Compared with PIFuHD, we get similar observations as in Sec.~\ref{sec:eval}, can generate high-quality shapes with less artifacts and more accurate estimated location (i.e. depth/global translation) 
than a normal-based one. 
Compared with PIFu, PaMIR and ICON, our SelfPIFu can recover significantly more details, and also more complete and reasonable shapes.
ECON also uses the depth map as the intermediate variable, and combined with SMPL to optimize a good geometric details, but the pose priority will introduces errors into the final results, especially the palm area, that cannot be ignored.

\section{Conclusion}
\label{sec:conclusion}
In this paper, we propose a novel self-supervised framework, named SelfPIFu, which is able to utilize abundant and diverse in-the-wild images lacking 3D GT during training, resulting in largely improved reconstructions when tested on unconstrained in-the-wild image.
Firstly, we empirically find that the estimated depth usually contains plausible details and is more robust than the estimated normal, and propose to use the depth map as the intermediate input for single-view 3D human reconstruction.
Then we further propose to exploit the inferred depth map as supervision to guide the learning of the implicit function by designing a novel depth-guided self-supervised SDF-based PIFu learning module.
This module contains two components, i.e. volume-aware and surface-aware supervision, to better utilize the depth information. 
As a result, our SelfPIFu improves the reconstruction quality compared with the normal-based PIfuHD and generalize better on real world images.
We believe that the SelfPIFu can lead to more accurate and robust reconstructions, particularly in real-world settings. 

\noindent\textbf{Limitation.} We utilize depth as an intermediate input in the overall SelfPIFu framework. 
Naturally, the accuracy of depth map affects the final results. 
Although the depth estimator predicts plausible and detailed results in standard and easy poses (e.g. standing-like poses), the robustness of the depth estimator it suffers from extreme poses.

\noindent\textbf{Future work.} 
The most straightforward approach to address extreme poses is to include such poses in the training data. However, given the challenges associated with acquiring extensive training data, an alternative approach is to explore the incorporation of SMPL priors into the depth estimator in future research.

\section{Acknowledgement} 
The work was supported by the Shenzhen General Project with No. JCYJ20220530143604010.

\bibliographystyle{splncs04}
\bibliography{cvmbib}
%
\end{document}